\documentclass[runningheads]{llncs}
\usepackage[T1]{fontenc}
\usepackage[skip=10pt]{caption} 
\usepackage{amsmath}
\usepackage{amssymb}
\usepackage{graphicx}
\usepackage{longtable}
\usepackage{enumitem}
\usepackage{booktabs}
\usepackage{afterpage}

\usepackage{lineno}



\begin{document}
\title{PEARL: Plan Exploration and Adaptive Reinforcement Learning for Multihop Tool Use}

\author{
Qihao Wang\inst{1,2} \and
Mingzhe Lu\inst{1} \and
Jiayue Wu\inst{1} \and
Yue Hu\inst{1}\thanks{Corresponding author. Email: huyue@iie.ac.cn} \and
Yanbing Liu\inst{1}
}

\institute{
1. Institute of Information Engineering, China \\
2. School of Cyber Security, University of Chinese Academy of Sciences, China
}

\maketitle              

\begin{abstract}
Large Language Models show great potential with external tools, but face significant challenges in complex, multi-turn tool invocation. They often exhibit weak planning, tool hallucination, erroneous parameter generation, and struggle with robust interaction. To tackle these issues, we present PEARL, a novel framework to enhance LLM planning and execution for sophisticated tool use. PEARL adopts a two-stage approach: an offline phase where the agent explores tools to learn valid usage patterns and failure conditions, and an online reinforcement learning phase. In the online phase, a dedicated Planner is trained via group Relative Policy Optimization (GRPO) with a carefully designed reward function that provides distinct signals for planning quality. Experiments on the ToolHop and T-Eval benchmarks show PEARL significantly outperforms existing methods, achieving a new state-of-the-art success rate of \textbf{56.5\%} on ToolHop while maintaining a low invocation error rate. Our work marks a key advance in addressing the complex planning challenges of tool use, contributing to the development of more robust and reliable LLM-based agents.
\end{abstract}


\section{Introduction}
\label{sec:introduction}

Large Language Models (LLMs) have demonstrated remarkable capabilities in natural language understanding and generation, marking a paradigm shift in artificial intelligence. However, their inherent knowledge is static and confined to their training data, limiting their ability to interact with the real world, access up-to-date information, or perform complex computations. A transformative solution to this limitation has emerged in augmenting LLMs with external tools \cite{patil2023gorilla,li2023api}. This approach, often termed tool learning, empowers LLMs to function as autonomous agents by granting them access to APIs, databases, and other specialized functions. Early successes, such as Toolformer \cite{schick2023toolformer}, showed that models could learn to make simple, single-step API calls, effectively extending their capabilities beyond the confines of text generation.

As the ambition for LLM agents grows, the focus is rapidly shifting from simple tool invocation to solving complex, multi-step problems that require the strategic orchestration of multiple tools in sequence \cite{kong2023tptu,gur2024real}. This leap in complexity exposes a critical bottleneck: the agent's capacity for coherent, long-horizon planning and robust execution. Current approaches often falter in this demanding arena. Many agents adopt a myopic, one-step-at-a-time strategy, failing to formulate or adhere to a long-term plan \cite{yao2022react}. This lack of foresight leads to brittle and unreliable execution, where agents are prone to hallucinating non-existent tools, generating faulty parameters, or failing to manage dependencies between consecutive tool calls \cite{wang2023survey}. Compounding this issue is a poor capacity for adaptation; when faced with an inevitable execution error, many agents cannot diagnose the failure, reflect on their plan, and dynamically adjust their strategy, often getting trapped in repetitive failure loops.

To address these fundamental challenges, we introduce \textbf{PEARL}: a framework for \textbf{P}lan \textbf{E}xploration and \textbf{A}daptive \textbf{R}einforcement \textbf{L}earning for Tool Use. PEARL tackles the intertwined problems of strategic planning and reliable execution through a principled, two-stage approach. First, to ensure robust execution, PEARL conducts an offline \textbf{tool exploration} phase. Here, the agent proactively interacts with each available tool in a controlled, trial-and-error manner to learn its valid usage patterns, argument formats, and failure conditions. This pre-emptive exploration builds a practical, learned user manual for each tool, drastically reducing execution errors. Second, to cultivate strategic thinking, PEARL employs a decoupled architecture where a dedicated \textbf{Planner} is trained using adaptive reinforcement learning. Optimized via Group Relative Policy Optimization (GRPO) \cite{shao2024deepseekmath}, this Planner learns to generate a complete, high-level plan before execution begins. The cornerstone of this training is a novel, \textbf{planning-centric reward function} that provides direct, dense feedback on the quality of the proposed tool chain, enabling the agent to effectively learn from both successes and failures and develop a genuine strategic capability.

We conduct extensive experiments on two challenging benchmarks: ToolHop and T-Eval. Our contributions are threefold:
\begin{itemize}
    \item We propose \textbf{PEARL},a novel framework that effectively integrates offline tool exploration for robust execution with online reinforcement learning for adaptive, strategic planning.
    \item We design an innovative planning-centric reward mechanism that makes the direct optimization of multi-step tool plans via RL both feasible and highly effective, guiding the agent toward optimal strategies.
    \item PEARL establishes a new \textbf{state-of-the-art}, achieving a \textbf{56.5\%} success rate on ToolHop—a significant absolute improvement over prior work—and demonstrating superior generalization on T-Eval, thereby validating the robustness and effectiveness of our approach.
\end{itemize}

\section{Related Work}
\subsection{LLM Tool Use}
The integration of tools with LLMs has evolved significantly from single-tool invocation to complex multi-tool workflows. Early approaches like Toolformer \cite{schick2023toolformer} demonstrated LLMs' ability to learn API calls through supervised fine-tuning on augmented text data. Subsequent work extended this paradigm: TALM \cite{parisi2022talm} explored tool-augmented learning, ToolkenGPT \cite{qian2024toolink} represented tools as learnable embeddings, and Gorilla \cite{patil2023gorilla} focused on retrieval-aware API generation. API-Bank \cite{li2023api} established critical benchmarks highlighting planning challenges in tool manipulation. Recent frameworks like HuggingGPT \cite{shen2023hugginggpt} and Visual ChatGPT \cite{wu2023visual} integrated multiple foundation models but struggled with long-horizon planning dependencies. While supervised methods \cite{liu2024toolace,shi2025tool} teach imitation of successful tool sequences, they remain brittle to novel scenarios and error recovery \cite{sun2024tools}. 

\subsection{Planning and Reasoning in LLMs}
Enhancing LLMs' planning capabilities has been approached through both prompting techniques and architectural innovations. Chain-of-Thought (CoT) prompting \cite{wei2022chain} pioneered explicit reasoning traces, while ReAct \cite{yao2022react} integrated reasoning with action execution for short-horizon tasks. Structured search methods like Tree-of-Thoughts (ToT) \cite{yao2023tree} and Graph-of-Thoughts (GoT) \cite{besta2023graph} enabled more sophisticated planning but required manual search configuration. Decoupled planning-execution architectures \cite{wang2023describe} improved modularity but lacked adaptive optimization. Embodied agent frameworks \cite{huang2023embodied,wang2024voyager} demonstrated planning in physical environments yet remained constrained by environment-specific designs. Similarly, prior efforts in tool planning \cite{wu2024toolplanner,liu2024tool} have been largely confined to imitation via fine-tuning, thereby forgoing the opportunity to refine plans with direct, step-wise feedback on tool selection.

\subsection{Reinforcement Learning for LLMs}
Reinforcement learning has proven effective for aligning LLMs with complex objectives. While RLHF \cite{ouyang2022training} optimized for human preferences, recent work adapts RL to specialized domains: RLTF \cite{liu2023rltf} for code generation, RRHF \cite{yuan2023rrhf} for response ranking, Tool,and RLP \cite{qiao2023making} for tool execution feedback. Tool learning frameworks like ToolStep \cite{yu2024steptool}  incorporated RL but focused primarily on execution rather than joint planning-execution optimization.  Our approach builds on this trend by using Group Relative Policy Optimization (GRPO) \cite{shao2024deepseekmath}, a robust PPO variant, to directly optimize the strategic quality of the plan, a distinct contribution from prior work that focused on execution-level rewards.

\section{Method}
\label{sec:method}

Our approach, PEARL, is designed to enhance the tool-using capabilities of LLMs by structuring the problem-solving process into a robust, two-stage framework. This framework first involves an offline exploration phase to master individual tool mechanics, followed by an online phase where a dedicated Planner, optimized via reinforcement learning, generates a strategic plan for execution. This decoupled approach allows us to tackle the challenges of both reliable execution and long-term planning systematically.

\begin{figure}[t]
    \centering
    \includegraphics[width=0.99\textwidth]{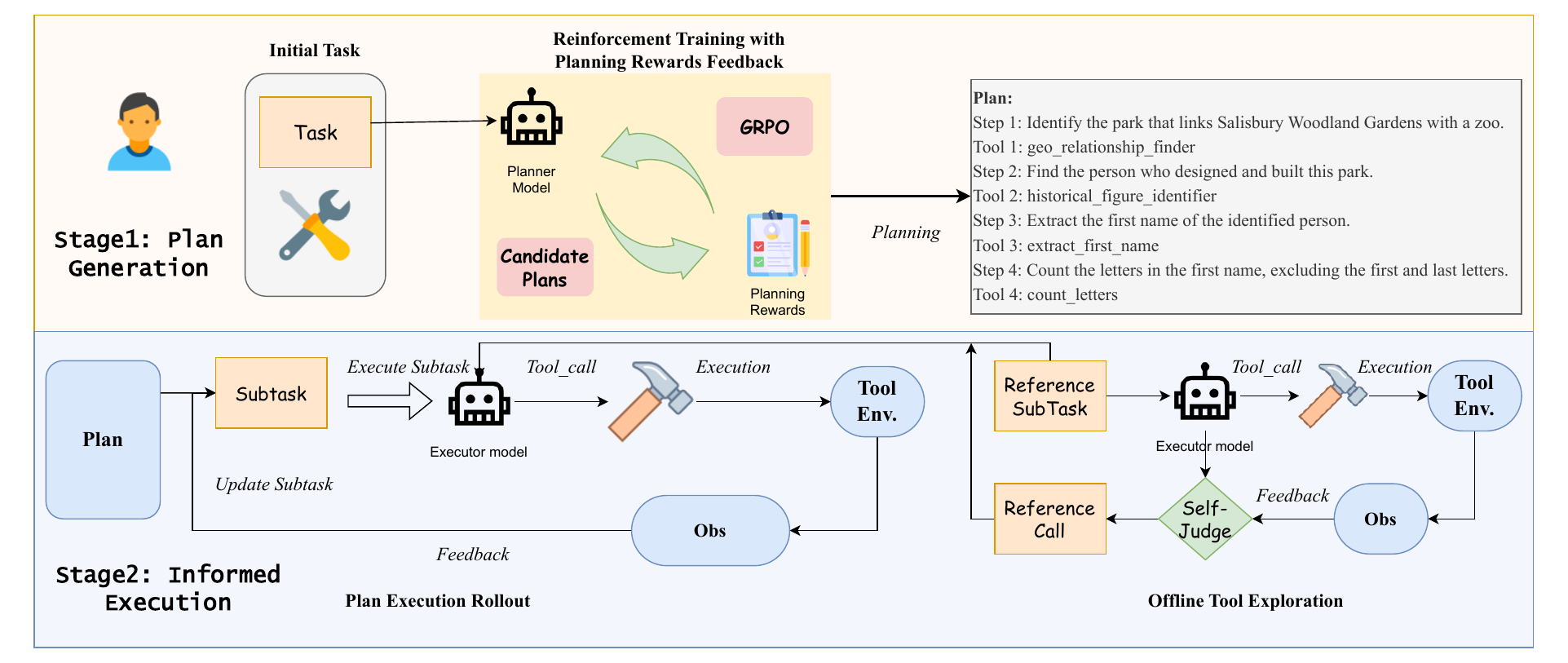}
    \caption{The PEARL framework's two-stage architecture. \textbf{Stage 1 (Plan Generation):} A dedicated Planner is trained via reinforcement learning (GRPO) with a planning-centric reward to create a multi-step plan. \textbf{Stage 2 (Informed Execution):} An Executor first learns tool usage through an offline exploration phase and then reliably executes the generated plan step-by-step.}
    \label{fig:framework}
\end{figure}
\subsection{Task Formulation}
Given a complex query $q$ and a collection of available tools $\mathcal{T} = \{t_1, t_2, \dots, t_k\}$, the agent's goal is to generate the correct final answer $a$. Each tool $t_j \in \mathcal{T}$ is defined by its name, a natural language description of its function, and the parameters it requires.

To solve the task, the agent must first formulate a multi-step plan $P$. This plan is an ordered sequence of sub-tasks:
\begin{equation}
    P = ((s_1, \tau_1), (s_2, \tau_2), \dots, (s_n, \tau_n))
\end{equation}
where $n$ is the number of steps in the generated plan. Each tuple $(s_i, \tau_i)$ represents the $i$-th sub-task, where $s_i$ is a natural language description of the step's goal (a sub-query), and $\tau_i \in \mathcal{T}$ is the specific tool selected to accomplish it. The overall process is a sequential decision-making problem where the output of one step can be essential for a subsequent step, demanding a coherent and logically sound plan.

\subsection{A Two-Stage Framework: Planning and Execution}
We propose a two-stage process: 1) Strategic Plan Generation, which is optimized with RL, and 2) Informed Execution, which is grounded by an offline exploration phase.

\subsubsection{Stage 1: Strategic Plan Generation}
In this stage, a dedicated \textbf{Planner} model, $\pi_{\theta}$, takes the initial user query $q$ as input and generates the entire multi-step plan $P$. The Planner's role is purely strategic; it must decompose the problem into a sequence of logical steps and select the appropriate tool for each step from the set $\mathcal{T}$. This process does not involve executing any tools. The output is the plan $P$, which acts as a blueprint for the subsequent execution stage. The core of our contribution lies in optimizing this Planner with reinforcement learning.

\subsubsection{Stage 2: Informed Execution}
The execution stage is handled by an \textbf{Executor} model, $\pi_{\phi}$, and is itself divided into two phases: an offline exploration phase to learn tool usage and an online phase to execute the given plan.

\paragraph{Offline Tool Exploration.} Before tackling any tasks, the agent proactively learns the semantics and syntax of each tool in an offline, trial-and-error process. By systematically attempting to call tools with various parameter configurations, the agent discovers valid invocation patterns, learns argument constraints, and identifies common failure modes. This exploration phase allows us to build a repository of successful tool call examples, effectively creating a practical, learned "user manual" for the toolset. This grounding step is crucial for mitigating tool hallucination and erroneous parameter generation during online execution. 

\paragraph{Online Step-by-Step Execution.} Once the Planner generates a plan $P$, the Executor model $\pi_{\phi}$ executes it step-by-step. For each step $i$ in the plan, the Executor receives a rich context $c_i$, including the original query $q$, the current sub-task $(s_i, \tau_i)$, and the history of execution results from previous steps, $H_{i-1} = \{o_1, \dots, o_{i-1}\}$. Crucially, the Executor leverages the knowledge acquired during the offline exploration phase to construct a valid tool call for $\tau_i$ with the correct parameters. This iterative process continues until all sub-tasks in the plan are completed, with the output of the final step, $o_n$, serving as the answer to the query.

\subsection*{Reward Design for Planning}
To optimize the Planner model $\pi_{\theta}$, we formulate a reward function that directly evaluates the quality of the generated plan $P$ referring to previous work in rewards design \cite{qian2025toolrl} . Our reward signal is designed to provide dense, step-by-step feedback on the correctness of the plan's structure. For a generated plan $P = ((s_1, \tau_1), \dots, (s_n, \tau_n))$, we compare it against a ground-truth plan $P^* = ((s^*_1, \tau^*_1), \dots, (s^*_m, \tau^*_m))$, where $m$ is the number of steps in the optimal plan.

The reward is calculated based on the correctness of the tool selection at each step. Specifically, for each step $i$ in the generated plan, we define a step-wise reward $r_i$:
\begin{equation}
  r_i = \left\{
  \begin{array}{ll}
    1/m & \quad \text{if } i \le m \text{ and } \tau_i = \tau^*_i \\
    0   & \quad \text{otherwise}
  \end{array}
  \right.
\end{equation}
This normalized reward provides a clear and well-scaled signal about whether the planned tool for a given step is correct. The total reward for the entire plan (trajectory) $P$ is the sum of the rewards for each of its $n$ steps:
\begin{equation}
    R(P) = \sum_{i=1}^{n} r_i
\end{equation}
This formulation ensures that a perfectly correct plan (where $n=m$ and all tools match) receives a total reward of 1. It naturally penalizes plans that are too short (by missing out on potential rewards) or too long (as extra steps beyond $m$ contribute zero reward), guiding the RL algorithm to learn policies that produce structurally and semantically correct tool plans of the appropriate length.

\subsection{Planner Optimization with GRPO}
To optimize the Planner model $\pi_{\theta}$ using the designed reward function, we employ Group Relative Policy Optimization (GRPO). GRPO is a variant of Proximal Policy Optimization (PPO) that enhances training stability by normalizing advantage estimates across a group of candidate plans generated for the same input query. This approach is particularly effective for our task, as it reduces the high variance often associated with RL-based fine-tuning of LLMs.

\paragraph{Group-wise Advantage Normalization.}
For each query $q$ in a training batch, we use the current policy $\pi_{\theta}$ to sample a group of $k$ distinct plans, $\mathcal{G}_q = \{P_1, \dots, P_k\}$. We calculate the total reward $R(P_j)$ for each plan in the group and then compute the group's reward mean $\mu_q$ and standard deviation $\sigma_q$. The advantage for each plan $P_j$ is then normalized:
\begin{equation}
    \hat{A}_j(P_j | q) = \frac{R(P_j) - \mu_q}{\sigma_q + \eta}
\end{equation}
where $\eta$ is a small constant for numerical stability. This normalization rescales rewards on a per-query basis, allowing the model to focus on the relative quality of different plans for the same problem.

\paragraph{Policy Optimization Objective.}
The policy $\pi_{\theta}$ is updated by maximizing the PPO-clip objective, which utilizes our group-normalized advantages. Let $\pi_{\text{old}}$ be the policy before the update. The probability ratio for a plan $P_j$ is $r_j(\theta) = \frac{\pi_\theta(P_j|q)}{\pi_{\text{old}}(P_j|q)}$. The optimization objective is then:
\begin{equation}
    \mathcal{L}_{\text{GRPO}}(\theta) = \mathbb{E}_{q, P_j \sim \pi_{\text{old}}} \left[ \min\left( r_j(\theta) \hat{A}_j, \text{clip}(r_j(\theta), 1-\epsilon, 1+\epsilon) \hat{A}_j \right) \right]
\end{equation}
where $\epsilon$ is the clipping hyperparameter. By optimizing this objective, the policy $\pi_\theta$ is encouraged to assign higher probabilities to plans that achieve above-average rewards for a given query, leading to a more robust and effective planning agent.

\section{Experiments}
\label{sec:experiments}

We conduct a comprehensive set of experiments to evaluate the effectiveness of our proposed framework, PEARL. Our evaluation is designed to answer three key questions: (1) Does PEARL outperform a wide range of state-of-the-art open-source and proprietary models on complex, multi-step tool-use tasks? (2) How generalizable is the planning capability learned by PEARL when paired with different execution models? (3) What is the contribution of each key component in our framework?

\subsection{Datasets and Environment}
We evaluate our method on two challenging benchmarks to assess the effectiveness of proposed method.
\begin{itemize}
    \item \textbf{ToolHop} \cite{ye2025toolhop}: A benchmark specifically designed for multi-hop tool reasoning. It requires agents to answer complex questions by composing a sequence of tool calls. We use the mandatory tool mode for our experiments.
    \item \textbf{T-Eval} \cite{chen2023t}:T-Eval is a benchmark for evaluating large language models' tool utilization capabilities by breaking down the process into sub-tasks like planning, reasoning, and review. From T-Eval's plan task, we've selected 100 instances and transformed them into executable Toolhop  style tasks.
\end{itemize}

\begin{table}[htbp]
\centering

\begin{tabular}{lccc}
\toprule
\textbf{Dataset} & \textbf{\# Tools} & \textbf{\# Instances} & \textbf{Avg. \# Steps} \\
\midrule
ToolHop          & 3912               & 995                 & 3.93 \\
T-Eval & 326            & 100                   & 3.26 \\
\bottomrule
\end{tabular}
\caption{Quantitative information of the datasets used in our evaluation.}
\label{tab:dataset_comparison}
\end{table}

\subsection{Experimental Settings}

\paragraph{Baselines.} We evaluate our method against a diverse set of 14 Large Language Models (LLMs) from four leading families, spanning both open-source and proprietary models of various scales. Additionally, we create a strong fine-tuned baseline to assess the benefits of standard task-specific optimization.

\begin{itemize}
    \item \textbf{Open-Source Models } \cite{2024qwen2.5}: Developed by Qwen, this series is known for its powerful multilingual performance and strong reasoning capabilities. We evaluate models of increasing scale to understand the impact of model size: \texttt{Qwen2.5-Instruct-7B}, \texttt{14B}, \texttt{32B}, and \texttt{72B}.

    \item \textbf{Proprietary Models }: We compare against some of the most capable and widely-used proprietary models. From Anthropic \cite{2024claude}, we test \texttt{Claude 3.5 Sonnet}. From OpenAI \cite{chatgpt2022}, we conduct experiments on \texttt{GPT-4o-mini} and \texttt{GPT-4o}. 

    \item \textbf{SFT-Optimized Model}: To establish a strong supervised learning baseline, we perform Supervised Fine-Tuning (SFT) on the \texttt{Qwen2.5-7B-Instruct} model using the training data from ToolHop. This baseline, denoted as \texttt{Qwen2.5-7B-SFT}, represents the standard approach for adapting a base model to a specific tool-use task.
\end{itemize}

\paragraph{Evaluation Metrics.}
We evaluate performance using two primary metrics:
\begin{itemize}
    \item \textbf{Success Rate (SR \%):} The primary metric, representing the percentage of tasks for which the agent produces the correct final answer.
    \item \textbf{Invocation Error Rate (IER \%):} The percentage of all attempted tool calls that result in an error in executing a whole task, either due to an invalid tool name (hallucination) or incorrect parameters. A lower rate indicates better execution reliability.
\end{itemize}

\subsection{Implementation Details}
Our PEARL framework and the SFT baseline are developed based on a 7B-parameter architecture (\texttt{Qwen2.5-7B}) to ensure a fair comparison against models of similar size. For PEARL, the Planner was trained using our Group Relative Policy Optimization algorithm for 15 epochs by VERL framework. The offline exploration phase was conducted for max of 10 rounds per tool to build the execution knowledge base. 

\subsection{Results and Discussion}

\subsubsection{Overall Performance}

Table \ref{tab:model_class_comparison} presents the main results on the ToolHop and T-Eval test sets. Our 7B model, PEARL, demonstrates remarkable performance, establishing a new state-of-the-art across the board and significantly outperforming models of much larger scale.

On the in-domain \textbf{ToolHop} benchmark, \textbf{PEARL-7B achieves a Success Rate of 56.5\%}, a substantial improvement over all other methods. Notably, it surpasses the specifically tuned \texttt{Qwen2.5-7B-SFT}  by over 10 absolute points, highlighting the superiority of our RL-based planning approach compared to standard imitation learning. More impressively, PEARL-7B also outperforms much larger models like \texttt{Qwen2.5-72B} and even the top-tier proprietary model \texttt{GPT-4o} . This result strongly suggests that PEARL's specialized architecture for planning and execution is more effective than simply scaling up model size for complex tool-use tasks. Furthermore, PEARL achieves a near-perfect \textbf{Invocation Error Rate of 3.8\%}, validating the effectiveness of our offline exploration phase in ensuring reliable tool calls.

On the generalization benchmark \textbf{T-Eval}, PEARL's strengths become even more apparent. It achieves an outstanding \textbf{Success Rate of 77.0\%}, not only dominating other open-source models but also surpassing the leading proprietary models, including \texttt{GPT-4o} and \texttt{Claude 3.5 Sonnet}. This exceptional generalization performance indicates that by learning the fundamental principles of planning via RL, PEARL develops a robust and transferable skill set that adapts seamlessly to unseen tools and tasks. Its IER of just \textbf{1.0\%} on this unseen dataset further underscores its reliability.

\begin{table}[h!]
\centering

\begin{tabular}{lcccc}
\toprule
\textbf{Model} & \multicolumn{2}{c}{\textbf{ToolHop }} & \multicolumn{2}{c}{\textbf{T-Eval }} \\
\cmidrule(lr){2-3} \cmidrule(lr){4-5}
& SR (\%) $\uparrow$ & IER (\%) $\downarrow$ & SR (\%) $\uparrow$ & IER (\%) $\downarrow$ \\
\midrule

\multicolumn{5}{l}{\textit{Base Open-Source Models}} \\
\midrule

\addlinespace[0.3em]
Qwen2.5-7B   &  9.8 & 28.8 & 25.0 & 29.0 \\
Qwen2.5-14B   & 26.3 & 15.7 & 34.0 & 14.0 \\
Qwen2.5-32B   & 25.0 & 12.4 & 26.0 & 9.0 \\
Qwen2.5-72B   & 45.5 & 13.2 & 54.0 & 9.0 \\

\midrule
\multicolumn{5}{l}{\textit{Proprietary Closed-Source Models}} \\
\midrule

\addlinespace[0.3em]
GPT-4o-mini & 40.2 & 11.6 & 68.0 & 10.0 \\
Claude 3.5 Sonnet & 39.9 & 19.6 & 71.0 & 16.0 \\
GPT-4o & 49.0 & 9.4 & 76.0 & 8.0 \\

\midrule
\multicolumn{5}{l}{\textit{Fine-Tuned Models}} \\
\midrule

\addlinespace[0.3em]
Qwen2.5-7B-SFT & 46.2 & 11.1 & 49.0 & 8.0 \\
PEARL-7B(Ours) & \textbf{56.5} & \textbf{3.8} & \textbf{77.0} & \textbf{1.0} \\

\bottomrule
\end{tabular}
\caption{Performance comparison across different model classes on the ToolHop and T-Eval benchmarks.}
\label{tab:model_class_comparison}

\end{table}

\subsubsection{Analysis of Planner Generalization}
A core hypothesis of our work is that a high-quality, generalizable planner is the key to success in complex tool-use. To isolate and verify the superiority of our PEARL-Planner, we conducted an experiment where we decoupled it from its native executor. We provided plans generated by our PEARL-Planner to other powerful models (\texttt{GPT-4o} and \texttt{Qwen2.5-72B}) and tasked them with executing the plan.

As shown in Table \ref{tab:plan_generalization}, the results are striking. When guided by the PEARL-Planner, the performance of both \texttt{GPT-4o} and \texttt{Qwen2.5-72B} improved dramatically. GPT-4o's Success Rate on ToolHop jumped from 49.0\% to 55.5\% (a 6.5\% absolute lift), while Qwen2.5-72B's SR surged from 45.5\% to 51.0\% (a 5.5\% absolute lift). This demonstrates that the plans generated by PEARL are not just effective but are of such high quality that they enable even the most powerful existing models to perform significantly better. This experiment confirms that the PEARL-Planner learns a universally effective planning strategy that transcends the specific executor it was trained with, showcasing its remarkable generalization capability.

\begin{table}[h!]
\centering

\begin{tabular}{lcc}
\toprule
\textbf{Model Configuration} & \textbf{SR (\%)} $\uparrow$ & \textbf{Success Lift} \\
\midrule
\textit{Baseline} \\
\quad Qwen2.5-72B  & 45.5 & - \\
\quad GPT-4o & 49.0 & - \\
\midrule
\textit{With PEARL-Planner Guidance} \\
\quad PEARL-Planner + Qwen2.5-72B Executor & 51.0 & \textbf{+5.5} \\
\quad PEARL-Planner + GPT-4o Executor & 55.5 & \textbf{+6.5} \\
\bottomrule
\end{tabular}
\caption{Analysis of planner generalization on the ToolHop benchmark.}
\label{tab:plan_generalization}

\end{table}

\subsubsection{Ablation Study}

To dissect the contribution of each component within PEARL, we conducted an ablation study on the ToolHop benchmark, with results presented in Table \ref{tab:ablation_study}.

\begin{table}[h!]
\centering
\begin{tabular}{lcc}
\toprule
\textbf{Method Variation} & \textbf{SR (\%)} $\uparrow$ & \textbf{IER (\%)} $\downarrow$ \\
\midrule
\textbf{PEARL (Full Model)} & \textbf{56.5} & \textbf{3.8} \\
\midrule
\multicolumn{3}{l}{\textit{Ablating RL Component:}} \\
\quad - w/o Planning Reward & 23.4 & 4.1 \\
\midrule
\multicolumn{3}{l}{\textit{Ablating Execution Component:}} \\
\quad - w/o Offline Exploration & 33.9 & 24.1 \\
\bottomrule
\end{tabular}
\caption{Ablation study of PEARL's core components on the ToolHop benchmark.}
\label{tab:ablation_study}
\end{table}

The results clearly validate our core design choices:

\begin{itemize}
    \item \textbf{Importance of Planning-Centric Reward:} This is the most critical component. Removing our specialized planning reward and relying only on a sparse final task success signal causes the Success Rate to plummet from 56.5\% to 23.4\%. This catastrophic drop highlights that our reward design is the cornerstone of the framework, effectively solving the credit assignment problem in long-horizon planning and guiding the agent to learn coherent, multi-step strategies.

    \item \textbf{Importance of Offline Exploration:} Removing the offline tool exploration phase, which pre-populates the executor's knowledge, leads to a severe degradation in reliability. The Invocation Error Rate spikes from a low 3.8\% to a staggering 24.1\%, with a corresponding drop in Success Rate to \textbf{33.9\%}. This directly demonstrates that preemptive, trial-and-error learning of tool mechanics is essential for achieving reliable execution and, consequently, high overall task success.
\end{itemize}
In summary, these ablations confirm that the synergistic combination of a sophisticated, reward-guided planner and a robust, exploration-hardened executor is indispensable to PEARL's state-of-the-art performance.

\section{Conclusion}
\label{sec:conclusion}

In this work, we introduced \textbf{PEARL}, a novel framework designed to systematically enhance the planning and execution capabilities of Large Language Models for complex, multi-step tool use. PEARL addresses critical limitations of existing methods, such as weak long-term planning and high error rates, through its unique two-stage architecture that decouples robust execution from adaptive, strategic planning. Our approach leverages offline tool exploration to build a reliable execution foundation, while an online reinforcement learning phase, guided by a carefully designed, planning-centric reward function, trains a dedicated Planner to master complex task decomposition. We validated our framework on two benchmarks. Our extensive experiments demonstrate that PEARL establishes new state-of-the-art performance, significantly outperforming not only standard fine-tuned models but also substantially larger open-source and leading proprietary models. Furthermore, our analysis revealed that the planning module of PEARL is highly generalizable, capable of elevating the performance of other powerful models, thereby validating the quality and universality of its learned strategies. This work represents a significant step towards developing more competent, reliable, and autonomous LLM-based agents, paving the way for their application in increasingly complex, real-world scenarios that demand strategic interaction with the digital world.

\bibliographystyle{splncs04}
\bibliography{mybibliography}
%

\end{document}